\documentclass[lettersize,journal]{IEEEtran}

\usepackage{graphicx} 
\usepackage{textcase} 
\usepackage{titlesec} 
\usepackage{xspace}
\usepackage{array}
\usepackage{framed}
\usepackage{subfigure}
\usepackage{enumitem}
\usepackage{pifont}
\usepackage{multirow}
\usepackage{amsmath} 
\usepackage{amssymb} 
\usepackage{float}
\usepackage{tcolorbox} 
\usepackage{amssymb} 
\newtcolorbox{answerBox}{
  colback=gray!20,   
  colframe=black,     
  coltitle=black,     
  boxrule=0.5pt,      
  arc=0pt,            
  left=2mm, right=2mm, top=2mm, bottom=2mm 
}

\usepackage{xurl} 
\usepackage{hyperref}

\usepackage{cite}

\newcommand{\tool}{\textsc{CopyCheck}\xspace}
\newcommand{\toolprob}{\textsc{Prob-Tool}\xspace}
\newcommand{\toolppl}{\textsc{Neighbor-PPL}\xspace}
\newcommand{\toolloss}{\textsc{Neighbor-LOSS}\xspace}

\definecolor{todocolor}{rgb}{0.9,0.1,0.1}

\title{As If We've Met Before: LLMs Exhibit Certainty in Recognizing Seen Files}

\author{
    \IEEEauthorblockN{
        Haodong Li\textsuperscript{1,*},
        Jingqi Zhang\textsuperscript{2,*},
        Xiao Cheng\textsuperscript{3},
        Peihua Mai\textsuperscript{2},
        Haoyu Wang\textsuperscript{1,**},
        Yan Pang\textsuperscript{2,**}
    }
    \thanks{* These authors contributed equally to this work.}
    \thanks{** Corresponding authors.}
    \\
    \IEEEauthorblockA{ 
        \textsuperscript{1}\textbf{Huazhong University of Science and Technology}\\
        Wuhan, China \\
        Email: \{lihd, haoyuwang\}@hust.edu.cn
    \\
        \textsuperscript{2}\textbf{National University of Singapore}\\
        Singapore\\
        Email: \{zhangjiggqi, peihua.mai\}@u.nus.edu, jamespang@nus.edu.sg
    \\
        \textsuperscript{3}\textbf{Macquarie University}\\
        Sydney, Australia \\
        Email: xiao.cheng@mq.edu.au
    }
}

\begin{document}

\maketitle
\begin{abstract}
The remarkable language ability of Large Language Models (LLMs) stems from extensive training on vast datasets, often including copyrighted material, which raises serious concerns about unauthorized use.
While Membership Inference Attacks (MIAs) offer potential solutions for detecting such violations, existing approaches face critical limitations and challenges due to LLMs' inherent overconfidence, limited access to ground truth training data, and reliance on empirically determined thresholds. 
 
We present \tool{}, a novel framework that leverages uncertainty signals to detect whether copyrighted content was used in LLM training sets. Our method turns LLM overconfidence from a limitation into an asset by capturing uncertainty patterns that reliably distinguish between ``seen" (training data) and ``unseen" (non-training data) content.
\tool{} further implements a two-fold strategy: (1) strategic segmentation of files into smaller snippets to reduce dependence on large-scale training data, and (2) uncertainty-guided unsupervised clustering to eliminate the need for empirically tuned thresholds.
Experiment results show that \tool{} achieves an average balanced accuracy of 90.1\% on LLaMA 7b and 91.6\% on LLaMA2 7b in detecting seen files.
Compared to the SOTA baseline, \tool{} achieves over 90\% relative improvement, reaching up to 93.8\% balanced accuracy.
It further exhibits strong generalizability across architectures, maintaining high performance on GPT-J 6B.
This work presents the first application of uncertainty for copyright detection in LLMs, offering practical tools for training data transparency.
\end{abstract}

\begin{IEEEkeywords}
LLM, uncertainty, MIA, copyright.
\end{IEEEkeywords}

 \vspace{-3mm}
\section{Introduction}

\IEEEPARstart{L}{arge} Language Models (LLMs) have emerged as powerful tools in natural language processing, excelling at tasks such as text summarization \cite{zhang2024benchmarking}, translation \cite{brants2007large}, and generation \cite{li2024pre}. The rise of commercial LLMs (e.g., ChatGPT \cite{OpenAI}) and open-source LLMs (e.g., LLaMA \cite{touvron2023llama}) has made these models widely accessible, allowing ever-growing users to benefit from applications such as chatbots for information retrieval and more. Their remarkable performance is highly related to billions of parameters, which require massive datasets during pre-training to capture diverse linguistic patterns and contextual richness \cite{villalobos2024will}.

The use of large-scale datasets for training LLMs raises significant concerns about copyright violations, as developers may include copyrighted materials without authorization.
For instance, Meta partially trained its LLM on the Books3 dataset, which contains over 170,000 copyrighted books \cite{books3news}. 
Major AI developers, including OpenAI, Meta, and Stability AI, are currently facing lawsuits over the unauthorized use of copyrighted works \cite{openai-sue, meta-sue, stable-sue}.
Partially due to these lawsuits, LLM developers are increasingly reluctant to disclose the specific details of the datasets used to train their models \cite{touvron2023llama, dubey2024llama, hurst2024gpt}. 
As a result, detecting unauthorized use of copyrighted materials in LLM training has become a critical research focus.

Membership Inference Attacks (MIAs) \cite{shokri2017membership, yeom2018privacy} have emerged as a powerful technique for identifying whether specific data, such as copyrighted works, were included in a model's training dataset. Initially developed for traditional machine learning models and classification tasks \cite{shokri2017membership, choquette2021label, rahman2018membership, long2020pragmatic}, some MIAs relied on training multiple shadow models to mimic the target model's behavior \cite{shokri2017membership, jayaraman2020revisiting}. However, this approach is impractical for LLMs because training even a single LLM requires massive computational resources (often hundreds or thousands of GPU hours) and billions of parameters, making it prohibitively expensive to train multiple shadow models.
Existing MIAs for LLMs can be broadly categorized into two main approaches. Reference-free methods \cite{yeom2018privacy, shi2023detecting, mattern2023membership, xie2024recall} use predefined metrics (e.g., perplexity or values derived from loss) and set threshold values to classify whether input data was part of the training set. Reference-based methods \cite{mireshghallah2022quantifying, mireshghallah2022empirical, fu2023practical} improve the selection of threshold by comparing the target model's behavior against reference models trained on datasets similar to the target model’s training dataset.

While most existing MIAs operate at the sentence level, copyright infringement detection often necessitates analyzing longer content such as books, leading to the development of \emph{file-level MIAs} \cite{meeus2024did}. These approaches \cite{shi2023detecting, meeus2024did, meeus2024copyright} are specifically engineered to process longer texts by incorporating their contextual characteristics. For instance, MIN-K\% PROB \cite{shi2023detecting} evaluates token probabilities across entire files, while the method proposed in \cite{meeus2024did} leverages statistical features of file-wide token probabilities to identify the membership status of books in the OpenLLaMA \cite{open_llama} model.
However, these file-level approaches face substantial limitations in practical implementation. A critical challenge stems from LLMs' tendency toward overconfidence~\cite{xiong2023can, Wang2024BLoBBL}, particularly when processing long files. This overconfidence manifests in LLMs making highly confident predictions, leading to unreliable outputs for MIAs and increased risk of membership misclassification.
The problem is particularly acute in scenarios with sparse ground truth data, where poor calibration further compounds the issue. For instance, loss-based MIAs \cite{shokri2017membership, yeom2018privacy, carlini2021extracting} rely heavily on token probabilities for loss computation. When these probabilities are distorted by overconfidence, the resulting membership signals become less reliable, significantly compromising the effectiveness of threshold-based detection methods.

To address this limitation, a more reliable indicator for membership detection is needed. Recent research~\cite{gawlikowski2023survey,kendall2017uncertainties, li2024malcertain} has demonstrated that \emph{uncertainty} metrics can effectively characterize model behavior and identify potential blind spots. Through empirical analysis (see \S\ref{study}), we have discovered a fundamental pattern: \emph{LLMs consistently exhibit lower uncertainty when processing content from their training data (i.e., seen content) compared to encountering unseen content}. This differential uncertainty pattern provides a robust signal for developing file-level MIAs that circumvent the challenges posed by model overconfidence.

However, implementing uncertainty-guided MIAs still presents two fundamental challenges:

\textbf{Challenge\#1: Scarcity of accessible ground truth membership labels.} Existing MIA methods typically require a substantial number of labeled samples—both members and non-members. However, acquiring such ground truth membership labels is often infeasible due to the opaque nature of LLM training datasets. Developers of state-of-the-art LLMs rarely disclose their pretraining data, citing copyright and competitive concerns. Consequently, only high-level descriptions of data sources are provided, while specific dataset details remain undisclosed~\cite{team2024gemini, team2023gemini, dubey2024llama, bai2023qwen}. This challenge is further exacerbated by recent events such as the shutdown of Books3—a widely used corpus for LLM training—following legal action by the anti-piracy group Rights Alliance~\cite{Anti-book3}.

\textbf{Challenge\#2: Reliance on empirically tuned thresholds.} Conventional MIA methods, including both reference-free and reference-based approaches, depend heavily on empirically determined thresholds that require access to ground truth membership labels for calibration. These thresholds are typically optimized on specific ground truth datasets and subsequently applied to target datasets under the assumption that both sets are identically distributed. In practice, however, this assumption rarely holds, and the limitation becomes particularly acute given the scarcity of accessible ground truth datasets highlighted in Challenge\#1.

To address these two fundamental challenges, we propose \tool{}, a novel file-level MIA framework for LLMs that \emph{transforms LLM overconfidence from a limitation into a methodological advantage} by leveraging uncertainty-based metrics to provide reliable indicators for membership detection. 
Our approach addresses \textbf{Challenge\#1} through a strategic file segmentation methodology that divides each file into smaller, analyzable snippets. This segmentation enables more granular uncertainty measurement while effectively augmenting the number of informative samples available for analysis. The snippet-level uncertainty signals are subsequently aggregated via max pooling to form a global uncertainty representation for the file.
To tackle \textbf{Challenge\#2}, we replace conventional threshold-based detection with unsupervised clustering algorithms, such as Gaussian Mixture Models (GMM)~\cite{reynolds2009gaussian}. This clustering-based approach adaptively differentiates between seen and unseen files based on their uncertainty representations without requiring manually tuned thresholds during inference, thereby enhancing both the robustness and generalizability of membership detection across diverse datasets and model configurations.

The main contributions of our work are outlined as follows:
\begin{itemize}
    \item To the best of our knowledge, this is the first work to apply uncertainty to copyright detection in LLMs. By transforming the inherent property of LLMs, i.e., overconfidence, from a limitation into an asset, our approach reveals its potential applications in file-level data privacy protection. 
    \item \tool{} eliminates the need for empirically tuned thresholds and large ground truth datasets through a two-fold solution that combines file segmentation with an uncertainty-based clustering mechanism.
    \item We evaluate \tool{} on real-world book datasets for seen file detection, achieving over 90\% average balanced accuracy on LLaMA 7B and LLaMA2 7B. It also generalizes well to GPT-J 6B, confirming its robustness across diverse LLM families.
\end{itemize}

 \vspace{-3mm}
\section{Background and Related Work}

\subsection{Membership Inference Attacks on LLMs}

Membership Inference Attacks (MIAs) aim to determine whether a specific data point was included in a model’s training set~\cite{shokri2017membership}. Originally designed for classification tasks in computer vision, MIAs have since been extended to language models~\cite{hisamoto2020membership,song2019auditing}. However, traditional shadow-model-based approaches~\cite{shokri2017membership} are impractical for LLMs due to their high computational cost and limited access to shadow data.

Existing MIAs for LLMs typically fall into two categories: (1) \textit{reference-free} methods that rely on metrics such as loss or perplexity~\cite{yeom2018privacy}, and (2) \textit{reference-based} methods that calibrate predictions using auxiliary models~\cite{mireshghallah2022quantifying,mireshghallah2022empirical}. Later works refined these strategies through difficulty calibration~\cite{watson2021importance}, likelihood ratio attacks (LiRA), or self-prompting to generate synthetic reference data~\cite{fu2023practical}. ReCaLL~\cite{xie2024recall} further leverages the ratio between conditional and unconditional likelihoods.

While these techniques are mostly applied at the sentence or sample level, recent copyright concerns in LLM pretraining—particularly involving books and other long-form content—have motivated the development of file-level MIAs~\cite{shi2023detecting, meeus2024did, meeus2024copyright}. For instance, \cite{shi2023detecting} builds the BookMIA benchmark and proposes a Min-K\% method that could detect excerpts from copyrighted books within GPT-3 \cite{brown2020language}’s pre-training data. \cite{meeus2024did} proposed a methodology for file-level membership inference on OpenLLaMA \cite{open_llama}, which collects member and non-member files (books and papers) and trains a meta-classifier using file-level features derived from token-level predictions. Unlike our work, both techniques suffer from the overconfidence problem and rely heavily on empirically determined thresholds or large collections of member and non-member files. In contrast, we propose a novel framework with uncertainty that addresses these challenges.

\vspace{-0.2in}
\subsection{Uncertainty Estimation in LLMs}

Although the capabilities of LLMs are widely recognized, their predictions are often regarded as overconfident~\cite{gawlikowski2023survey,xiong2023can,Wang2024BLoBBL}, especially after fine-tuning~\cite{balabanov2024uncertainty,wang2023lora}.
To better quantify this behavior, a growing body of work has focused on estimating uncertainty in LLM outputs.

Uncertainty estimation methods for LLMs can be categorized into single-LLM methods, sampling-based methods, and ensemble-based methods, with the latter two leveraging interactions with open-source LLMs to improve estimation.

\noindent \textbf{Single-LLM Methods.} 
For a single open-source LLM, the model typically generates a deterministic output for each input. To estimate uncertainty under this setting, Kuhn et al.~\cite{Kuhn2023SemanticUL} introduce semantic entropy to estimate uncertainty in question answering tasks.
Lin et al.~\cite{Lin2022TeachingMT} fine-tune the LLM on a dataset that includes confidence scores, thereby enhancing the LLM's ability to express uncertainty. For a single closed-source LLM, Xiong et al. ~\cite{xiong2023can} explore several strategies to stimulate the expression of uncertainty in LLM, including using prompt engineering to directly query uncertainty, quantifying the differences in Top-K~\cite{tian-etal-2023-just} predictions as a measure of uncertainty, and adopting reasoning-based techniques such as Chain of Thought (CoT)~\cite{Kojima2022LargeLM} and Multi-Step methods to encourage LLM to express uncertainty.

\noindent \textbf{Sampling-based Methods.}
Bayesian methods differ from traditional single-LLM approaches: rather than relying on fixed model parameters, Bayesian LLMs sample parameters from a distribution during each prediction. This leads to varying probability predictions for the same input, capturing the model’s uncertainty~\cite{gawlikowski2023survey,118274,10.1145/168304.168306}. For instance, Wang et al.\cite{Wang2024BLoBBL} introduced the Bayesian LoRA (BLOB) method, where the low-rank matrix $A$ in LoRA is treated as a Bayesian matrix, with its parameters sampled during each prediction in the fine-tuning stage. Furthermore, a simple approach is to use Monte Carlo Dropout (MCD) to approximate complex Bayesian variational inference~\cite{Gal2015DropoutAA,Gal2015BayesianCN}. By enabling dropout during inference, MCD simulates different network configurations through multiple samplings, effectively incorporating uncertainty and achieving a similar effect to Bayesian inference.

\noindent \textbf{Ensemble-based Methods.}
Ensemble-based methods take a broader approach by combining multiple LLMs rather than relying on a single model. By constructing an ensemble of LLMs, each with subtle differences, a set of predictions can be produced for a given sample, with the variations between these predictions serving as a measure of uncertainty. These differences can be introduced in several ways, such as using LLMs with different parameter scales~\cite{carlini2021extracting}, employing different initialization parameters before fine-tuning~\cite{lakshminarayanan2017simple,Balabanov2024UncertaintyQI}, or training with varying numbers of epochs~\cite{li2024malcertain}.

Single-LLM methods, although simple and computationally efficient, often provide inaccurate uncertainty estimates and tend to underestimate the true uncertainty of LLMs. In contrast, sampling-based methods offer more precise uncertainty estimation but come with the drawback of requiring multiple inferences, resulting in higher computational overhead. Ensemble-based methods, by combining multiple sub-LLMs, enhance the reliability of uncertainty estimation but incur significant computational and storage costs. In our work, we focus on sampling-based methods and ensemble-based methods to achieve more reliable uncertainty estimation. 

 \vspace{-3mm}
\section{Uncertainty Applications in LLMs: An Empirical Analysis}
\label{study}
 \vspace{-1mm}
In this section, we conduct an empirical analysis to explore how LLM uncertainty can be applied to detect seen files within a dataset containing a mix of seen and unseen files. 
To this end, we fine-tune the target LLM on a dataset containing both seen and unseen samples to obtain an uncertainty estimation LLM, and then estimate uncertainty for each sample in this fine-tuning dataset. 
Next, we perform separate statistical analyses on the uncertainty of these seen and unseen samples, and compare the results. 
We aim to investigate \emph{whether we can effectively leverage uncertainty to distinguish between samples seen and unseen by the target LLM.}

\begin{figure*}[!t]
\centering
 \vspace{-4mm}
\includegraphics[width=0.9\textwidth]{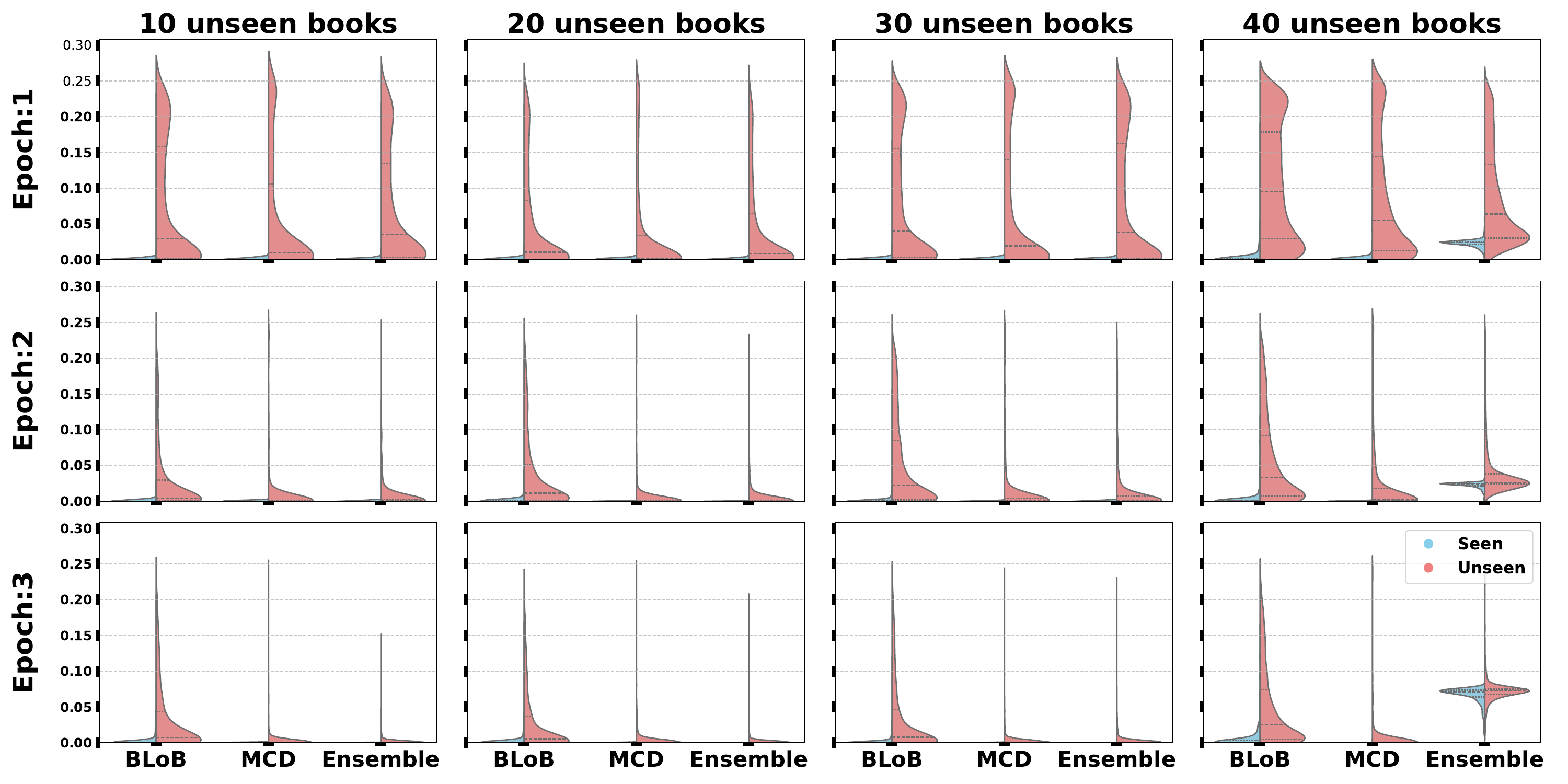}
 \vspace{-3mm}
\caption{Violin plots of aleatoric uncertainty across different uncertainty estimation methods (BLoB, MCD, and Ensemble), varying numbers of unseen books $N_\text{unseen} \in \{10, 20, 30, 40\}$, and fine-tuning epochs ($1$, $2$, and $3$), for seen (blue) and unseen (red) snippets in the Suspected Seen Dataset.}
\label{fig_violin}
 \vspace{-4mm}
\end{figure*}

 \vspace{-3mm}
\subsection{Problem Modeling}

We model \textit{seen file detection} as a binary classification task, where files seen by the target LLM are labeled as 1, and unseen files are labeled as 0.
In practical scenarios, we are often provided with a collection of files—referred to as the \textit{Suspected Seen Dataset}—which comprises a mixture of both seen and unseen files, although the exact composition is unknown. We aim to accurately distinguish between these two categories within this dataset. 
To facilitate this, we assign all files in the \textit{Suspected Seen Dataset} an initial label of 1. We then collect an equally sized dataset of new reliably unseen files, termed the \textit{Unseen Dataset}, and label these files as 0. Notably, the files labeled as ``unseen'' in the dataset are considered reliable, as they are collected from files that appeared after the release of the target LLM.

 \vspace{-2mm}
\subsection{Empirical Analysis Setup}
\subsubsection{Dataset} 
\label{study_dataset}
Our dataset construction relies on the widely accepted understanding that collecting unseen samples is generally easier than collecting seen samples, as the former can be sourced from materials published after the LLM’s release date~\cite{Li2024DiggerDC,shi2023detecting}.

\noindent \textbf{Files Collection.} We use the BookMIA dataset~\cite{shi2023detecting}, which consists of 100 books curated for benchmarking MIA methods. This dataset focuses on detecting pretraining data used by OpenAI models prior to 2023. Within the 100 books, 50 books published after 2023 are reliably labeled as ``unseen", while the other 50 books are labeled as ``seen". To expand our collection of unseen data, we collaborated with industry partners to manually collect an additional 100 books published after January 2023. This brings the total to 150 unseen books and 50 seen books, where each book is treated as a single file.

\noindent \textbf{Dataset Construction.} 
The dataset consists of 100 books. Specifically, we randomly select 50 books from the 150 unseen books to form the \textit{Unseen Dataset}. Then, we create the \textit{Suspected Seen Dataset} by combining $N_\text{unseen}$ books from the remaining 100 unseen books with $50 - N_\text{unseen}$ randomly selected seen books from the BookMIA dataset. We vary $N_{\text{unseen}} \in \{10, 20, 30, 40\}$ to simulate different proportions of unseen books within the \textit{Suspected Seen Dataset}. Each book is divided into snippets of 512 words, yielding approximately 100 snippets per book. After the above process, the final dataset contains approximately 5000 snippets labeled as 0 (from the \textit{Unseen Dataset}) and 5000 snippets labeled as 1 (from the \textit{Suspected Seen Dataset}).

\subsubsection{Uncertainty Estimation and Quantification}

We employ three widely used uncertainty estimation methods for LLMs: Bayesian Low-Rank Adaptation (BLoB)\cite{Wang2024BLoBBL}, Monte Carlo Dropout (MCD), and Ensemble\cite{lakshminarayanan2017simple,Balabanov2024UncertaintyQI}. First, these methods fine-tune the target LLM to construct an uncertainty estimation LLM. Then, this model generates a set of prediction probability distributions for each snippet in the dataset. Finally, the variation among these probability distributions is used as a measure of uncertainty, where greater variation corresponds to higher uncertainty. Detailed descriptions of these uncertainty estimation methods are provided in \S\ref{uc_llm}.

To further clarify the source of the probability distributions, we process the LLM output by extracting only the logit values corresponding to the tokens ``0" and ``1"~\cite{Wang2024BLoBBL}. These logits are then transformed into a two-element probability distribution using the Softmax function, ensuring that the probabilities sum to 1. This distribution represents the LLM’s confidence in classifying the snippet as either ``0" or ``1". To measure uncertainty, we adopt aleatoric uncertainty, a widely used metric in uncertainty quantification where details is in \S\ref{metrics}.

\subsubsection{Results Analysis}

To assess the effectiveness of leveraging uncertainty in distinguishing between seen and unseen samples, we divide the snippets in the Suspected Seen Dataset into two groups: those that are truly seen and those that are truly unseen. We then measure their Aleatoric Uncertainty, with the results presented in the Violin Plots in Figure~\ref{fig_violin}.

We conduct experiments across three factors: the uncertainty estimation methods (BLoB, MCD, and Ensemble), the number of unseen books $N_\text{unseen}$ in the Suspected Seen Dataset ($N_\text{unseen}=10, 20, 30, \text{and } 40$), and the number of fine-tuning epochs ($1, 2, \text{and } 3$). 

For $N_\text{unseen}=10$, the Aleatoric Uncertainty of seen snippets using these three uncertainty estimation methods is nearly 0 after fine-tuning for 1 epoch, while unseen snippets exhibit significantly higher Aleatoric Uncertainty, with some values exceeding 0.2. 
As $N_\text{unseen}$ increases, both BLoB and MCD show no significant change in performance when distinguishing between the two groups of snippets, whereas Ensemble's performance deteriorates. For example, when $N_\text{unseen}=40$, using Ensemble for fine-tuning 1 epoch, the Aleatoric Uncertainty for all seen snippets is below 0.03, while about 50\% of unseen snippets have their uncertainty concentrated around 0.03.
As the number of fine-tuning epochs increases, the distinction between the two groups of snippets diminishes. For $N_\text{unseen}=40$, after fine-tuning for 3 epochs, the Aleatoric Uncertainty distributions of the two groups using the Ensemble method become nearly identical.

\begin{answerBox}
\vspace{-1mm}
\textbf{Findings:}
All three uncertainty estimation methods show strong effectiveness in distinguishing between seen and unseen snippets, regardless of the value of $N_\text{unseen}$. Among them, BLoB consistently outperforms both MCD and Ensemble, with the most effective distinction achieved after fine-tuning for a single epoch.
\vspace{-1mm}
\end{answerBox}

 \vspace{-3mm}
\section{Overview of \tool{}}
\label{method}
\begin{figure*}[!t]
\centering
 \vspace{-4mm}
\includegraphics[width=0.9\textwidth]{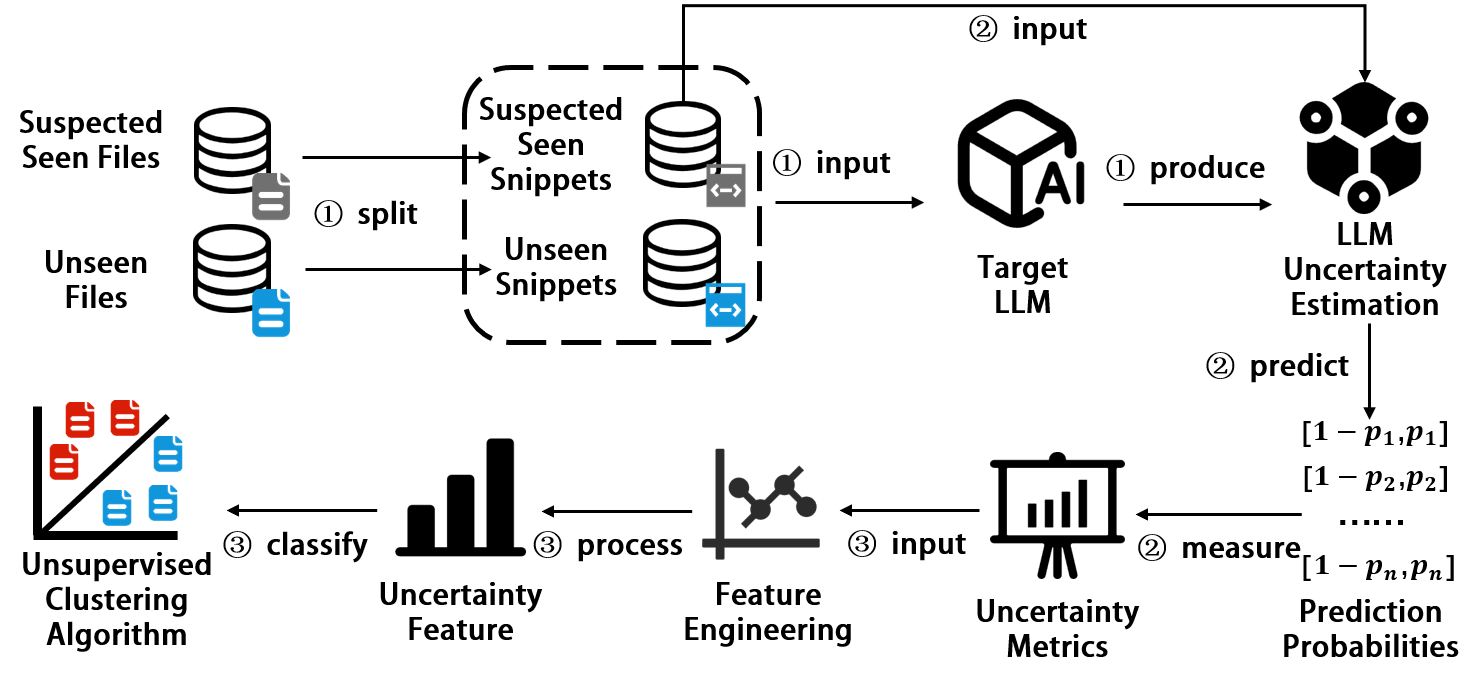}
 \vspace{-4mm}
\caption{Framework of \tool{}. This framework contains three main phases, namely, \ding{172} Uncertainty Estimation for LLMs, \ding{173} Uncertainty Metrics Quantification, and \ding{174} Seen File Detection. Gray represents files and snippets suspected to have been seen by the target LLM, red represents those that have been seen, and blue represents those that have not been seen.}
 \vspace{-5mm}
\label{fig_framework}
\end{figure*}

Based on the Key Findings in \S ~\ref{study}, we propose ~\tool{}, a framework designed to effectively detect seen files from the Suspected Seen Dataset. The framework is illustrated in Figure~\ref{fig_framework}. Specifically, \tool{} consists of three main phases: 

\ding{172} Uncertainty Estimation for LLMs: For a given set of suspected seen files, we initially assign a pseudo label of 1 to each file. Next, we collect an equal number of files published after the release date of the target LLM and label them as 0. Each file is split into equally sized snippets, which are then used to fine-tune the target LLM using three uncertainty estimation methods (BLoB, MCD, and Ensemble), with each method producing a distinct uncertainty estimation LLM.

\ding{173} Uncertainty Metrics Quantification: For each snippet in the Suspected Seen Dataset, we apply three LLMs with uncertainty estimation capabilities to generate their respective sets of prediction probabilities. For each set of prediction probabilities, we compute three types of uncertainty metrics, resulting in 13 distinct metrics per set. This yields a total of $3 \times 13=39$ comprehensive metrics, providing a thorough uncertainty estimation for each snippet.

\ding{174} Seen File Detection: To detect seen files, we first apply Principal Component Analysis (PCA)~\cite{abdi2010principal} to the $39$-dimensional uncertainty metrics of each snippet, extracting the most significant features into a $10$-dimensional vector. Then, we apply a max pooling strategy across all snippets in a file to aggregate the most informative signals, resulting in a single $10$-dimensional uncertainty representation for each file. Finally, these file-level vectors are fed into an unsupervised clustering algorithm, which separates the files into two groups. The group exhibiting lower uncertainty is treated as ``seen", while the other cluster is treated as ``unseen".

\vspace{-3mm}
\subsection{Uncertainty Estimation for LLMs}

This section describes the process of estimating uncertainty for LLMs, which serves as an indicator for membership inference attack (i.e., seen file detection) via unsupervised clustering, as detailed in Section~\ref{sec:seen_file_detection}. 
To achieve this goal, we employ different strategies depending on the accessibility of the target LLM. For white-box LLMs, we fine-tune the model to enable uncertainty estimation, rather than directly classifying whether a file was included in the pre-training dataset.

We first construct a dataset by augmenting the Suspected Seen Dataset (labeled as 1) with unseen files (labeled as 0).
This dataset is then used to fine-tune the target model $M$ using an uncertainty estimation method, yielding the uncertainty estimation model $M'$, which is then applied to perform inference on the Suspected Seen Dataset.

In \tool{}, this fine-tuning setup naturally introduces a differential exposure between the two file types: seen files appear in both the pre-training and fine-tuning phases, while unseen files are introduced only during fine-tuning. This exposure difference allows $M'$ to capture meaningful uncertainty signals that reliably indicate which files were likely present in $M$’s pre-training data.
The detailed setup is following:

\subsubsection{Dataset Construction.}
For a given set of suspected seen files, which are initially assigned a pseudo label of 1, we aim to determine whether they have been unseen by $M$. Next, an equal number of files that appeared only after the release of $M$ are collected as unseen files and labeled them as 0. Each file is divided into $N$ snippets based on its length $l$, with each snippet inheriting the label of its parent file.

\subsubsection{Uncertainty Estimation Methods} 
\label{uc_llm}
We employ three widely used uncertainty estimation methods for LLM, with each method producing a distinct uncertainty estimation LLM $M'$. These methods include two sampling-based approaches, Bayesian Low-Rank Adaptation (BLoB)~\cite{Wang2024BLoBBL} and Monte Carlo Dropout (MCD), as well as an ensemble-based method, Ensemble~\cite{lakshminarayanan2017simple,Balabanov2024UncertaintyQI}.

\begin{itemize}[noitemsep, topsep=1pt, partopsep=1pt, listparindent=\parindent, leftmargin=*]

\item \textbf{Bayesian Low-Rank Adaptation (BLoB).}
The BLoB method~\cite{Wang2024BLoBBL} constructs the low-rank matrix  $A$ of LoRA \cite{hu2021lora} as a Bayesian-parameterized matrix, where the parameters are treated as distributions instead of fixed deterministic values. During the fine-tuning phase, the weights are modeled as following a low-rank Gaussian prior, with the mean set to the pre-trained weights and the covariance structure defined by a low-rank matrix. The posterior distribution is updated iteratively through learning on the fine-tuning dataset. During inference, parameters are sampled from this posterior to generate stochastic predictions.

\item \textbf{Monte Carlo Dropout (MCD).} During the fine-tuning phase, MCD follows standard LoRA fine-tuning procedure, utilizing dropout as a regularization technique to mitigate overfitting. In the inference phase, the dropout layers remain active, allowing the LLM to perform multiple forward passes for a given input. These repeated predictions are treated as samples drawn from an implicit posterior distribution, enabling uncertainty estimation.

\item \textbf{Ensemble.} The Ensemble method involves fine-tuning the target LLM multiple times with different random seeds, forming a collection of sub-models that produce slightly varied predictions for the same input.
\end{itemize}

Following previous works~\cite{li2024malcertain,Wang2024BLoBBL}, we perform 10 sampling iterations for each sampling-based method, and fine-tune 10 sub-LLMs for the Ensemble method.

 \vspace{-4mm}
\subsection{Uncertainty Metrics Quantification}
\label{metrics}
After fine-tuning $M$ to obtain the uncertainty estimation LLM $M'$, we estimate uncertainty for each snippet in the Suspected Seen Dataset. Specifically, for a given snippet, we input it into $M'$  to generate a set of predictions. 
For a generative LLM, its output is a probability distribution over the entire vocabulary, where each element represents the probability of a specific token being generated at the current step. To adapt this output for binary classification, we perform a pruning operation~\cite{Wang2024BLoBBL} on each output vector, retaining only the probabilities corresponding to the tokens ``0" and ``1". These tokens are pre-defined to represent the two possible classes in our work.
Next, the softmax function is applied to this two-element pruned vector to normalize the probabilities, ensuring that they sum to 1. The $i$-th output generated by $M'$ for a given snippet is represented as $\hat{p_i} = [1 - p_i, p_i]^T$, where $p_i$ denotes the probability assigned to the token ``1". For each snippet, $n$ such outputs are generated. In our experiments, an uncertainty estimation LLM can
generate 10 $(n=10)$ prediction probabilities for each snippet.

To comprehensively estimate the target LLM uncertainty for suspected seen snippets, we go beyond the conventional measures of Aleatoric and Epistemic uncertainty~\cite{gawlikowski2023survey,helton1996guest,hullermeier2021aleatoric}. In addition, we introduce a novel discrepancy-based measure between the prediction probability $\hat{p_i}$ and the label $y$, which we term Factual uncertainty. The details of each uncertainty metric are as follows: 

\noindent \textbf{Aleatoric Uncertainty.} This uncertainty, also referred to as ``data uncertainty", arises from the inherent complexity of the data itself and reflects variability within individual prediction distributions that cannot be reduced through learning. 
Specifically, when an LLM has been exposed to a set of samples, it tends to be more familiar with previously encountered samples compared to unseen ones, thus reducing the uncertainty linked to data complexity. As a result, the LLM exhibits lower aleatoric uncertainty for samples it has already seen.
Following prior work\cite{kwon2018uncertainty,shridhar2019comprehensive}, we compute it as:

\[
U_{\text{Aleatoric}} = \frac{1}{n} \sum_{i=1}^n \Big( \text{diag}(\hat{p}_i) - \hat{p}_i \hat{p}_i^T \Big)
\]

\noindent \textbf{Epistemic Uncertainty.} 
Also referred to as ``model uncertainty", this arises from the LLM's insufficient knowledge or incomplete understanding of certain information. For a given sample, uncertainty estimation LLM generates a set of predictive probabilities, measuring the variance among these predictions. Once the model is exposed to a specific sample, it typically accumulates more information, thereby enhancing its understanding of the sample and enabling more confident and accurate predictions. Consequently, epistemic uncertainty for these samples decreases. Following prior work\cite{kwon2018uncertainty,shridhar2019comprehensive}, we compute it as:

\[
U_{\text{Epistemic}} = \frac{1}{n} \sum_{i=1}^n \Big( (\hat{p}_i - \bar{p})(\hat{p}_i - \bar{p})^T \Big),
\]
where $\bar{p} = \frac{1}{n} \sum_{i=1}^n \hat{p}_i$.

To further measure the differences among this set of prediction probabilities, we simplify each prediction to \( p_i \), which represents the LLM's predicted probability for ``1", as the predicted probability for ``0" is linearly dependent. Consequently, we abstract this set of predictions as $P = \{p_1, p_2, \dots, p_n\}$. Following prior work~\cite{li2024malcertain,10.1145/3377811.3380368,li2021can}, we also select the following six metrics to quantify the differences within the set of predictions: 

Predictive Entropy~\cite{bein2006entropy} is computed as the negative sum of the products of each probability and the logarithm of that probability:
\[
H(P) = -\sum_{i=1}^n p_i \log(p_i).
\]

Standard Deviation~\cite{altman2005standard} is calculated as the square root of the mean squared difference between each probability $p_i$ and the mean $\bar{P}$ of $P$:

\[
\sigma(P) = \sqrt{\frac{1}{n} \sum_{i=1}^n (p_i - \bar{P})^2},
\]
where $\bar{P} = \frac{1}{n} \sum_{i=1}^n p_i$.

Kullback-Leibler Divergence~\cite{kullback1951kullback} is calculated as  the sum of the products of each probability in the target distribution $P$ and the logarithm of the ratio of $P$ to the reference distribution $Q = \{\bar{P}, \bar{P}, \dots, \bar{P}\}_n$:

\[
D_{\text{EU-KL}}(P \| Q) = \sum_{i=1}^n p_i \log\left(\frac{p_i}{\bar{P}}\right)\]


$\Delta Max(P)$ is calculated as the difference between the largest value and the second-largest value in $P$:
\[
\Delta Max(P) = \max(P) - \text{second\_max}(P).
\]

$\Delta Min(P)$ is calculated as the difference between the second-smallest value and the smallest value in $P$:
\[
\Delta Min(P) = \text{second\_min}(P) - \min(P).
\]

$\Delta MM(P)$ is calculated as the difference between the mean and the median of $P$:
\[
\Delta MM(P) = \bar{P} - \text{median}(P).
\]

\noindent \textbf{Factual Uncertainty.} It refers to the difference between the prediction made by the uncertainty estimation LLM and the label $y$. These differences can provide insights into the LLM’s understanding of the label. Specifically, in the fine-tuning dataset, there is a large number of samples that the target LLM has not seen before, which are labeled as 0. However, for unseen samples labeled as 1, significant discrepancies between the predicted values and the labels arise due to inconsistencies in the fine-tuning dataset’s annotations. To quantify these differences, we employ five widely used disparity metrics:

Manhattan distance~\cite{malkauthekar2013analysis} is calculated as the sum of the absolute differences between corresponding elements of two probability distributions $P$ and $Y=\{y, y, \dots, y\}_n$:
\[
D_{\text{Manhattan}} = \sum_{i=1}^n |p_i - y|.
\]

Euclidean distance~\cite{malkauthekar2013analysis} is calculated as the square root of the sum of squared differences between corresponding elements of two probability distributions $P$ and $Y$:
\[
D_{\text{Euclidean}} = \sqrt{\sum_{i=1}^n (p_i - y)^2}.
\]

Negative Log-Likelihood (NLL)~\cite{bosman2000negative} is calculated as the negative sum of the product of the true labels and the logarithm of the predicted probabilities for each class:
\[
\text{NLL} = -\sum_{i=1}^n y \log(p_i).
\]

Chebyshev distance~\cite{coghetto2016chebyshev} is calculated as the maximum absolute difference between corresponding elements of two probability distributions $P$ and $Y$:
\[
D_{\text{Chebyshev}} = \max_{i=1, 2, \dots, n} |p_i - y|.
\]

Kullback-Leibler Divergence~\cite{kullback1951kullback} is calculated as:
\[
D_{\text{FU-KL}}(P \| Y) = \sum_{i=1}^n p_i \log\left(\frac{p_i}{y}\right).
\]

In summary, for each given snippet, an uncertainty estimation LLM generates a set of predicted probabilities, from which 13 distinct uncertainty metrics can be derived. In our framework, we utilize three such LLMs, resulting in a total of 39 uncertainty metrics per snippet.

\vspace{-3mm}
\subsection{Seen File Detection\label{sec:seen_file_detection}}
In this section, we focus on constructing a global representation of file-level uncertainty and utilize it to perform unsupervised clustering on the Suspected Seen Dataset. Specifically, we standardize the snippet-level uncertainty metrics and apply Principal Component Analysis (PCA) for dimensionality reduction, resulting in a 10-dimensional vector per snippet. Subsequently, we utilize max pooling to aggregate the most informative features across all snippets, yielding a concise 10-dimensional uncertainty vector that characterizes each file. These vectors serve as input for the final clustering using an unsupervised learning algorithm. The detailed process is as follows:

\subsubsection{Feature Engineering}

Feature standardization is a critical preprocessing step in feature engineering. It transforms each feature to have a mean of 0 and a standard deviation of 1 using the formula:
\[
    z = \frac{x - \mu}{\sigma}.
\]
We standardize 39 uncertainty metrics per snippet to ensure that all features are on a comparable scale, which facilitates fair and effective model training.

Subsequently, we apply Principal Component Analysis (PCA)~\cite{mackiewicz1993principal} to compress the standardized features. PCA is a classical unsupervised dimensionality reduction technique that transforms the original high-dimensional features into a set of orthogonal components. These components are ranked according to the amount of variance they capture, selecting the top components that retain the most critical information while discarding redundant dimensions. In our work, the 39 original uncertainty metrics are compressed into 10 principal components.

After constructing snippet-level uncertainty features, we further obtain a global uncertainty representation for each file by applying a max-pooling operation. Specifically, given the 10-dimensional uncertainty feature of all snippets within a file, the maximum value for each dimension is extracted to form a new 10-dimensional uncertainty feature vector.

\subsubsection{Unsupervised Clustering Algorithms}
\label{clustering_algo}
We rely on unsupervised clustering algorithms to classify the suspected seen files.
Based on data similarity, these algorithms divides the files into two clusters. The one with higher uncertainty is treated as representing unseen files, and the one with lower uncertainty as seen files.
In this work, we apply three widely used unsupervised clustering algorithms: 
\begin{itemize}[noitemsep, topsep=1pt, partopsep=1pt, listparindent=\parindent, leftmargin=*]

\item \textbf{Gaussian Mixture Model (GMM)}~\cite{reynolds2009gaussian}. GMM models the data as a mixture of multiple Gaussian distributions, estimating the parameters of these distributions by maximizing the likelihood. In our work, we use GMM to partition the files into two Gaussian distributions, representing the distributions of seen and unseen files.

\item \textbf{Hierarchical Clustering (HC)}~\cite{murtagh2012algorithms}. HC is a widely used unsupervised clustering algorithms that builds a tree-like structure (dendrogram) to represent the hierarchical relationships within the data. In our work, we use a bottom-up agglomerative hierarchical clustering method. Data similarity is calculated using Euclidean distance, which measures the straight-line distance between points in multidimensional space. For cluster merging, we apply Ward’s method, which merges the pair of clusters that results in the smallest increase in within-cluster variance.

\item \textbf{K-Means}~\cite{ahmed2020k}.
K-Means is a clustering algorithm that partitions data points into $K$ clusters, calculates the centroid of each cluster, and iteratively refines the cluster assignments. In our work, we set $K=2$ to partition the files into two clusters, representing seen and unseen files.
\end{itemize}

 \vspace{-2mm}
\section{Evaluation}
\label{evaluation}

We explore the effectiveness of \tool{} in detecting seen files from the Suspected Seen Dataset by seeking answers to the following three research questions:
\begin{itemize}[noitemsep, topsep=1pt, partopsep=1pt, listparindent=\parindent, leftmargin=*]
\item \textbf{RQ1:} How effective is \tool{} in detecting seen files?
\item \textbf{RQ2:} Can \tool{} outperform the state-of-the-art tool?
\item \textbf{RQ3:} Can \tool{} generalize effectively to different data sources and LLMs with different architectures?
\item \textbf{RQ4:} How effective is \tool{} in black-box LLM settings?
\end{itemize}

 \vspace{-4mm}
\subsection{Evaluation Setup}
\vspace{-1mm}
We implement \tool{} using the Pytorch framework and two GeForce A100 40GB GPUs driven by CUDA.

\subsubsection{Dataset}

Our source data includes 200 books collected from two sources. The first source is the BookMIA~\cite{shi2023detecting} dataset, which includes 50 books seen by the target LLM and 50 books unseen by it. The second source comprises 100 books published after January 2023, which we manually collected in collaboration with our industry partners to serve as unseen by the target LLM. Consequently, our source data includes 50 books considered seen and 150 considered unseen by the target LLM.

We construct four datasets, each containing 100 books, with varying numbers of books seen by the target LLM set to 10, 20, 30, and 40, respectively. The construction of these datasets is described in \S\ref{study_dataset}.

\subsubsection{Target LLMs} 
In RQ1 and RQ3, we select three widely used open-source LLMs that meet the requirements of the BookMIA dataset: LLaMA 7B, LLaMA2 7B, and GPT-J 6B.
All models are publicly available on Hugging Face~\cite{member} under the repositories 
``huggyllama/llama-7b", ``meta-llama/Llama-2-7b", and ``EleutherAI/gpt-j-6b".
In RQ2, we select the same LLMs used in ~\cite{meeus2024did}, namely OpenLLaMA 3B and OpenLLaMA 7B. These models are also available on Hugging Face under the repositories ``openlm-research/open\_llama\_3b" and ``openlm-research/open\_llama\_7b".

\subsubsection{Baseline Approaches}

To demonstrate the effectiveness of \tool{}, we conduct a comparative evaluation against four state-of-the-art approaches. An overview of these baselines is provided in Table \ref{tab:eval-targets}, detailing their underlying methodologies.

\begin{table}[htbp]
\centering
    \caption{Details of baseline approaches.
    }
    \vspace{-3mm}
    \resizebox{0.45\textwidth}{!}{
   \begin{tabular}{|>{\centering\arraybackslash}m{2cm}|>{\centering\arraybackslash}m{5cm}|}
        \hline 
         \textbf{Approach} & \textbf{Description} \\
         \hline 
         \hline
    
        \toolprob ~\cite{shi2023detecting}
        &Computes the mean of the bottom K\% token probabilities as an estimate of prior exposure.
        \\ \cline{1-2}

        \toolloss ~\cite{mattern2023membership}
        &Measures the relative deviation of a sample's loss from its nearest neighbours to detect exposure.
        \\ \cline{1-2}
        
        \toolppl ~\cite{mattern2023membership}
        &Same as \toolloss, but uses perplexity to measure deviation. 
        \\ \cline{1-2}
        
        Meeus et al.~\cite{meeus2024did}
        &Queries the LLM for token-level probabilities, converting them into handcrafted features for a supervised meta-classifier. Requires substantial labeled data for training. Finally, an unsupervised clustering algorithm is employed to discriminate between seen and unseen files.        
        \\ 
        \hline
    \end{tabular}
    }
     \vspace{-3mm}
    \label{tab:eval-targets}
\end{table}

\subsubsection{Evaluation Metrics} 

We evaluate the performance of \tool{} using standard classification metrics: True Positives (TP), False Positives (FP), False Negatives (FN), and True Negatives (TN).

\noindent
\textbf{True Positives (TP)}: Number of files correctly identified as \textit{seen}. \\
\textbf{False Positives (FP)}: Number of files incorrectly identified as \textit{seen} when they are actually \textit{unseen}. \\
\textbf{False Negatives (FN)}: Number of files incorrectly identified as \textit{unseen} when they are actually \textit{seen}. \\
\textbf{True Negatives (TN)}: Number of files correctly identified as \textit{unseen}.

Using these values, we calculate two key metrics. First, \textbf{accuracy (Acc.)} measures the overall proportion of correctly identified files:

\[\text{Accuracy} = \frac{TP + TN}{TP + TN + FP + FN}.\]

Second, to account for potential class imbalance, we report \textbf{balanced accuracy (bAcc.)}, which is the average of the True Positive Rate (TPR) and True Negative Rate (TNR):
\[
\text{Balanced Accuracy} = \frac{\text{TPR} + \text{TNR}}{2},
\]
where $\text{TPR} = \frac{TP}{TP + FN}$ and $\text{TNR} = \frac{TN}{TN + FP}$.

\vspace{-2mm}
\subsection{RQ1: Effectiveness of Seen File Detection}
\label{rq1}

In RQ1, we first evaluate the effectiveness of \tool{} in seen file detection, followed by two ablation studies to assess the contributions of uncertainty features and the unsupervised clustering algorithm.

\subsubsection{Seen File Detection}
Table~\ref{anomaly_detection_result_llama} summarizes the results of \tool{} on detecting seen files under different settings of $N_\text{unseen}$. For LLaMA 7B and LLaMA2 7B, \tool{} maintains high and consistent performance across the three clustering algorithms (GMM, HC, and K-Means). When $N_\text{unseen}=10$, it achieves accuracies of 92.0\% and 94.0\%, respectively, with balanced accuracies above 87.5\%. As $N_\text{unseen}$ increases to 20, performance slightly improves, reaching up to 95.8\% balanced accuracy, while a mild decline is observed at $N_\text{unseen}=40$ due to class imbalance between seen and unseen files. Across all settings, both false positives (FP) and false negatives (FN) remain low, with only a few seen files misidentified as unseen and vice versa. Overall, \tool{} demonstrates robust detection of seen files across all configurations, confirming its reliability within the Suspected Seen Dataset.

\renewcommand{\arraystretch}{1.1}
\begin{table*}[h]
    \centering
     \vspace{-3mm}
    \caption{Using 50 suspected seen books, evaluation of \tool{} for detecting \textbf{seen} files by LLaMA 7b and LLaMA2 7b.}
     \vspace{-2mm}
    \resizebox{0.9\linewidth}{!}{
    \begin{tabular}{c|c|cc|cccc|cc|cccc}
    \hline
     \multicolumn{2}{c|}{BookMIA}  &\multicolumn{6}{c|}{LLaMA 7B} 
     &\multicolumn{6}{c}{LLaMA2 7B} \\ \hline
   Tool &$N_{\text{unseen}}$ &Acc.(\%) &bAcc.(\%) &TP&FP&FN&TN &Acc.(\%) &bAcc.(\%)  &TP&FP&FN&TN\\
    \hline \hline

   \toolprob& \multirow{9}{*}{10} &28.0 &55.0 &4 &0 &36 &10 &28.0 &51.3 &5 &1&35 &9 \\
    \cline{1-1}   \cline{3-14} 

\toolppl& &52.0 &62.5  &18  &2  &22  &8 &44.0  &57.5  &14&2&26&8 \\
    \cline{1-1}   \cline{3-14} 

\toolloss&  &52.0 &62.5  &18  &2  &22  &8 &38.0&53.8&11&2&29&8   \\
    \cline{1-1}   \cline{3-14} 
    
    IF&&80.0 &68.8 &35 &5 &5 &5 &84.0 &75.0 &36 &4 &4 &6 \\
    OCSVM&&80.0 &65.0 &36 &4 &6 &4 &70.0 &55.0 &32 &7 &8 &3 \\
    DBSCAN&&70.0 &81.3 &25 &0 &15 &10 &56.0 &72.5 &18 &0 &22 &10 \\
    \cline{1-1} \cline{3-14}
    \textbf{\tool{} (GMM)}& & \textbf{92.0} & \textbf{87.5} &38 &2 &2 &8 &\textbf{94.0} &\textbf{92.5} &38 &1 &2 &9\\ 
    \textbf{\tool{} (HC)}&& \textbf{92.0}&\textbf{87.5} &38 &2 &2 &8 &\textbf{94.0} &\textbf{92.5} &38 &1 &2 &9 \\
    \textbf{\tool{} (K-Means)}&& \textbf{92.0} &\textbf{87.5} &38 &2 &2 &8 &\textbf{94.0} &\textbf{92.5} &38 &1 &2 &9 \\
    
    \hline \hline

    \toolprob& \multirow{9}{*}{20} &56.0 &46.7 &28 &20 &2 &0 &44.0 &53.3 &2 &0 &28 &20 \\
    \cline{1-1} \cline{3-14}

\toolppl& &48.0 &54.2 &7  &3  &23  &17  &56.0  &54.2  &19  &11  &11  &9 \\
    \cline{1-1}   \cline{3-14} 

\toolloss& &42.0 &51.7 &1  &0  &29  &20  &40.0  &50.0  &0  &0  &30  &20 \\
    \cline{1-1}   \cline{3-14} 
    
    IF&&64.0 &58.3 &26 &14 &4 &6 &68.0 &62.5 &27 &13 &3 &7 \\
    OCSVM&&58.0 &54.2 &26 &13 &4 &7 &52.0 &54.2 &13 &7 &17 &13 \\
    DBSCAN&&80.0 &83.3 &20 &0 &10 &20 &68.0 &73.3 &14 &0 &16 &10 \\
    \cline{1-1} \cline{3-14} 
    \textbf{\tool{} (GMM)}& &92.0 &92.5 &27 &1 &3 &19 &\textbf{96.0} &\textbf{95.8} &29 &1 &1 &19\\ 
    \textbf{\tool{} (HC)}&&\textbf{94.0} &\textbf{93.3} &29 &2 &1 &18 &94.0 &94.2 &28 &1 &2 &19 \\
    \textbf{\tool{} (K-Means)}&&92.0 &92.5 &27 &1 &3 &19 &94.0 &94.2 &28 &1 &2 &19 \\
   
    \hline \hline
    
    \toolprob& \multirow{9}{*}{30} &40.0 &45.8 &15 &25 &5 &5 &72.0 &65.0 &6 &0 &14 &30 \\
    \cline{1-1} \cline{3-14} 

\toolppl& &58.0  &57.5  &11  &12  &9  &8   &56.0  &57.5  &13  &15  &7&15 \\
    \cline{1-1}   \cline{3-14} 

\toolloss&&54.0 &53.3  &10  &13  &10  &17   &44.0  &50.8  &17  &25  &3&5  \\
    \cline{1-1}   \cline{3-14} 
    
    IF&&44.0 &50.0 &16 &24 &4 &6 &52.0 &58.3 &18 &22 &2 &8 \\
    OCSVM&&64.0 &69.2 &19 &17 &1 &13 &50.0 &55.0 &16 &21 &4 &9 \\
    DBSCAN&&86.0 &82.5 &13 &0 &7 &30 &76.0 &70.0 &8 &0 &12 &30 \\
    \cline{1-1} \cline{3-14}
    \textbf{\tool{} (GMM)}& &\textbf{94.0} &\textbf{92.5} &17 &0 &3 &30 &96.0 &95.8 &19 &1 &1 &29 \\ 
    \textbf{\tool{} (HC)}&&92.0 &90.8 &17 &1 &3 &29 &\textbf{98.0} &\textbf{97.5} &19 &0 &1 &30 \\
    \textbf{\tool{} (K-Means)}&&\textbf{94.0} &\textbf{92.5} &17 &0 &3 &30 &96.0 &96.7 &20 &2 &0 &28 \\
   
    \hline \hline

    \toolprob& \multirow{9}{*}{40} &40.0 &55.0 &8 &28 &2 &12 &36.0 &48.8 &7 &29 &3 &11 \\
    \cline{1-1} \cline{3-14}

\toolppl& &80.0 &50.0 &0  &0  &10  &40 &80.0 &50.0 &0  &0  &10  &40 \\
    \cline{1-1}   \cline{3-14} 

\toolloss& &80.0 &50.0 &0  &0  &10  &40  &20.0  &46.3  &9  &39  &1  &1 \\
    \cline{1-1}   \cline{3-14} 
    
    IF&&32.0 &50.0 &8 &32 &2 &8 &32.0 &50.0 &8 &32 &2 &8 \\
    OCSVM&&42.0 &52.5 &7 &26 &3 &14 &44.0 &57.5 &8 &26 &2 &14 \\
    DBSCAN&&80.0 &50.0 &0 &0 &10 &40 &80.0 &50.0 &0 &0 &10 &40 \\
    \cline{1-1} \cline{3-14}
    \textbf{\tool{} (GMM)}&&\textbf{94.0} &\textbf{88.8} &8 &1 &2 &39 &\textbf{84.0} &\textbf{82.5} &8 &6 &2 &34 \\
    \textbf{\tool{} (HC)}&&\textbf{94.0} &\textbf{88.8} &8 &1 &2 &39 &\textbf{84.0} &\textbf{82.5} &8 &6 &2 &34 \\
    \textbf{\tool{} (K-Means)}&&86.0 &87.5 &9 &6 &1 &34 &\textbf{84.0} &\textbf{82.5} &8 &6 &2 &34 \\
    
    \hline
    \end{tabular}
    }
     \vspace{-4mm}
    \label{anomaly_detection_result_llama}
\end{table*}

\subsubsection{Probability Features vs. Uncertainty Features}
Probability-based features are widely used in MIA to represent the model’s confidence for a given input. To evaluate the role of uncertainty features in \tool{}, we conduct an ablation study comparing three probability-based features with our proposed uncertainty features in detecting seen files. Specifically, we include (1) the Min-K\% Prob method proposed by Shi et al.~\cite{shi2023detecting}, which averages the bottom K\% token probabilities to estimate prior exposure, and (2) the Neighbour Attack method~\cite{mattern2023membership}, which measures the relative deviation of a sample’s loss or perplexity from its nearest neighbours in the representation space.


To construct a global probability feature vector for each file, we first extract ten Min-K\% probability values under different K settings and compute ten different loss and perplexity values of ten neighbors for each snippet in the Suspected Seen Dataset, resulting in a set of initial snippet-level probability feature vectors. We then apply the same feature engineering pipeline as \tool{} to aggregate these snippet-level vectors into a global file-level representation. Finally, these file-level vectors are fed into a GMM to perform clustering, classifying the files as either seen or unseen.
The experiment results are presented in Table ~\ref{anomaly_detection_result_llama} under the rows labeled "\toolprob”, "\toolppl” and "\toolloss”. 

Specifically, all probability-based baselines perform notably worse than \tool{}. When $N_\text{unseen}=10$, \toolprob achieves only 28.0\% accuracy and 55.0\% balanced accuracy on LLaMA 7B, correctly identifying a few unseen files (TN=10) but misclassifying most seen ones (FN=36). Neighbour-PPL and Neighbour-LOSS perform slightly better, reaching 52.0\% accuracy and 62.5\% balanced accuracy, yet still fail to distinguish seen and unseen files effectively. Even under their best settings—\toolprob at $N_\text{unseen}=30$ (72.0\% accuracy, 65.0\% bAcc) and Neighbour-based features at $N_\text{unseen}=40$ (up to 80.0\% accuracy but only 50.0\% bAcc)—their performance remains far below that of \tool{}, whose lowest accuracy is 84.0\% and balanced accuracy 82.5\%.  

Overall, \tool{} achieves average balanced accuracies of 90.1\% on LLaMA 7B and 91.6\% on LLaMA2 7B, compared to only 50.6\%–54.6\% for the probability-based baselines. These results highlight the substantial advantage of uncertainty-based features in capturing exposure signals.

\subsubsection{Clustering vs. Anomaly Detection}

Next, we conduct an ablation experiment to evaluate the contributions of anomaly detection and clustering algorithms to the overall performance of \tool{}. 

In addition to the three unsupervised clustering algorithms introduced in \S\ref{clustering_algo}, we also employ three unsupervised anomaly detection algorithms to detect seen files. These algorithms approach anomaly detection from different perspectives, such as density, distance, and isolation, to identify samples with high uncertainty that may deviate from the overall data distribution. Specifically, we employ three unsupervised anomaly detection algorithms:

\begin{itemize}[noitemsep, topsep=1pt, partopsep=1pt, listparindent=\parindent, leftmargin=*]

\item \textbf{DBSCAN}~\cite{khan2014dbscan}. It is a density-based anomaly detection algorithm that identifies clusters by detecting regions with high data density. Points located in sparse areas, which do not fit well into any cluster, are considered anomalies.

\item \textbf{Isolation Forest (IF)}~\cite{liu2008isolation}. 
It identifies anomalies by constructing multiple random trees that ``isolate''  data points. Anomalies, which are easier to isolate, can be detected by measuring how frequently a point is isolated across the trees. The more frequently a data point is isolated, the more likely it is to be an anomaly.

\item \textbf{One-Class SVM (OCSVM)}~\cite{manevitz2001one}. It is a kernel-based anomaly detection algorithm that learns a decision boundary around the majority of the data points. It identifies anomalies as points that lie outside this learned boundary. By maximizing the margin in a high-dimensional feature space, OCSVM effectively separates normal instances from potential outliers, treating the latter as deviations from the learned data distribution.

\end{itemize}

We categorize the files into normal and anomalous groups using the anomaly detection algorithms, treating the group with higher average uncertainty as unseen files. The rows labeled ``IF", ``OCSVM", and ``DBSCAN" in Table ~\ref{anomaly_detection_result_llama} present the results. Compared to \tool{} with clustering algorithms, \tool{} with anomaly detection algorithms consistently underperforms in seen file detection. Under LLaMA 7B with $N_\text{unseen}=10$, \tool{} achieves balanced accuracies of 68.8\%, 65.0\%, and 81.3\% with IF, OCSVM, and DBSCAN, respectively—substantially lower than the 90.1\% achieved by clustering. A similar performance gap is observed in other configurations. For instance, when $N_\text{unseen}=20$, the best-performing anomaly detection algorithm (DBSCAN) still trails the clustering algorithm by over 20\% in balanced accuracy for LLaMA2 7B. These results suggest that while anomaly detection provides a reasonable lens for modeling high-uncertainty outliers, it lacks the precision of clustering algorithms in separating seen and unseen samples. This underscores the critical role of clustering in the overall performance of \tool{}.

\begin{answerBox}
 \vspace{-1mm}
\textbf{Answer to RQ1:} The experiment results demonstrate that \tool{} is effective in detecting unseen files, achieving average balanced accuracies of 90.1\% and 91.6\% on LLaMA 7B and LLaMA2 7B, respectively, evaluated on 50 books with varying ratios of unseen books.
Ablation experiments further emphasize the importance of uncertainty features and unsupervised clustering.
 \vspace{-1mm}
\end{answerBox}

\vspace{-4mm}
\subsection{RQ2: \tool{} vs. The SOTA Tool}
\vspace{-1mm}
To demonstrate the effectiveness of \tool{}, we compare it against state-of-the-art (SOTA) tool~\cite{meeus2024did} in file-level MIA. The SOTA tool queries the target LLM to obtain token-level probabilities, which are then aggregated into handcrafted statistical features. These features serve as inputs to a supervised meta-classifier designed for membership inference. However, it requires a substantial amount of labeled training data to function effectively.

\begin{table*}[ht]
    \centering
    \vspace{-2mm}
    \caption{Using 50 suspected seen books, evaluation of \tool{} for detecting seen files by Open LLaMA 3B and 7B.}
     \vspace{-2mm}
    \resizebox{0.9\linewidth}{!}{
    \begin{tabular}{c|c|cc|cccc|cc|cccc}
    \hline
    \multicolumn{2}{c|}{BookMIA} & \multicolumn{6}{c|}{Open LLaMA 3B} & \multicolumn{6}{c}{Open LLaMA 7B} \\ \hline
    Tool & $N_{\text{unseen}}$ & Acc.(\%) & bAcc.(\%) & TP & FP & FN & TN & Acc.(\%) & bAcc.(\%) & TP & FP & FN & TN \\
    \hline\hline
    Meeus et al.~\cite{meeus2024did} & \multirow{2}{*}{10} & 52.0 & 51.3 & 21 & 5 & 19 & 5 & 54.0 & 52.5 & 22 & 5 & 18 & 5 \\
    \textbf{\tool{}}                 &                      & \textbf{90.0} & \textbf{90.0} & 36 & 1 & 4 & 9 & \textbf{92.0} & \textbf{91.3} & 37 & 1 & 3 & 9 \\
    \cline{1-1}\cline{2-14}
    Meeus et al.~\cite{meeus2024did} & \multirow{2}{*}{20} & 50.0 & 49.2 & 16 & 11 & 14 & 9 & 54.0 & 53.3 & 17 & 10 & 13 & 10 \\
    \textbf{\tool{}}                 &                      & \textbf{96.0} & \textbf{96.7} & 28 & 0 & 2 & 20 & \textbf{100.0} & \textbf{100.0} & 30 & 0 & 0 & 20 \\
    \cline{1-1}\cline{2-14}
    Meeus et al.~\cite{meeus2024did} & \multirow{2}{*}{30} & 48.0 & 48.3 & 10 & 16 & 10 & 14 & 40.0 & 40.0 & 8 & 18 & 12 & 12 \\
    \textbf{\tool{}}                 &                      & \textbf{96.0} & \textbf{95.0} & 18 & 0 & 2 & 30 & \textbf{96.0} & \textbf{95.0} & 18 & 0 & 2 & 30 \\
    \cline{1-1}\cline{2-14}
    Meeus et al.~\cite{meeus2024did} & \multirow{2}{*}{40} & 48.0 & 48.8 & 5 & 21 & 5 & 19 & 48.0 & 48.8 & 5 & 21 & 5 & 19 \\
    \textbf{\tool{}}                 &                      & \textbf{96.0} & \textbf{93.8} & 9 & 1 & 1 & 39 & \textbf{90.0} & \textbf{86.3} & 8 & 3 & 2 & 37 \\
    \hline
    \end{tabular}
    }
    \label{sota_tool}
    \vspace{-4mm}
\end{table*}

We reproduce the experimental setup described in Meeus et al.~\cite{meeus2024did}. Specifically, we train the meta-classifier using the Random Forest algorithm, following their configuration. For training data, we randomly select 1,000 seen books from ~\cite{member}, and 1,000 unseen books from ~\cite{non-member}. Both the SOTA tool and \tool{} (with GMM clustering algorithm) are evaluated on the four datasets described in \S\ref{study_dataset}.
Table \ref{sota_tool} summarizes the comparison between \tool{} and the SOTA method on Open LLaMA 3B and 7B. Across all $N_\text{unseen}$ settings, \tool{} achieves 90–100\% accuracy and above 93\% balanced accuracy, whereas the baseline remains near chance level (about 50\%). The most significant gap appears at $N_\text{unseen}=10$, where \tool{} surpasses the SOTA method by over 70 points in balanced accuracy. The margin remains stable across different model scales, confirming the robustness of our uncertainty-driven approach.

\begin{answerBox}
\vspace{-1mm}
\textbf{Answer to RQ2:} These experiment results demonstrate that \tool{} consistently outperforms the SOTA tool, achieving over 90\% relative improvement in balanced accuracy.
\vspace{-1mm}
\end{answerBox}

\vspace{-3mm}
\subsection{RQ3: Generalizability of \tool{}}

\textbf{Different LLM Architectures.} In RQ1, we assess the effectiveness of \tool{} in identifying files seen by two LLMs based on the LLaMA architecture. To further assess the generalizability of \tool{}, we select another LLM based on the Transformer architecture, GPT-J 6B, and evaluate the performance of \tool{} (using the GMM algorithm) on the same four datasets described in \S\ref{study_dataset}.

\begin{table}[h]
    \centering
     \vspace{-2mm}
    \caption{Evaluating the effectiveness of \tool{} for GPT-J 6B.}
     \vspace{-2mm}
    \resizebox{0.9\linewidth}{!}{
    \begin{tabular}{c|c|c|cccc}
    \hline
     BookMIA & \multicolumn{6}{c}{GPT-J 6B} \\ \hline
    \multicolumn{1}{c|}{$N_{\text{unseen}}$} & Acc.(\%) & bAcc.(\%) & TP & FP & FN & TN \\
    \hline\hline
    10 & 86.0 & 80.0 & 36 & 3 & 4 & 7 \\
    20 & 98.0 & 98.3 & 29 & 0 & 1 & 20 \\
    30 & 94.0 & 93.3 & 18 & 1 & 2 & 29 \\
    40 & 92.0 & 83.8 & 7  & 1 & 3 & 39 \\
    \hline
    \end{tabular}
    }
     \vspace{-3mm}
    \label{gptj_6b}
\end{table}

Table~\ref{gptj_6b} shows that \tool{} achieves comparable performance to that on LLaMA models (see Table~\ref{sota_tool}), with an average accuracy of 92.5\% and balanced accuracy of 88.9\% across the four datasets. The highest performance occurs at $N_\text{unseen}=20$, reaching 98.0\% accuracy and 98.3\% balanced accuracy, where all unseen files are correctly identified and only one seen file is misclassified.
The results confirm that \tool{} maintains strong detection capability even on LLMs with different architectures, indicating robust generalizability.

\textbf{Different Data Sources.} To further assess \tool{}'s performance across diverse data sources, we conduct additional experiments using the ArxivMIA dataset~\cite{liu2024probing}. We randomly select 500 papers from ArxivMIA and partition them into two groups based on publication date: papers published after January 2023 are treated as seen files, while those published before January 2023 are treated as unseen files. The parameter $N_\text{unseen}$ denotes the number of unseen papers within the suspected seen dataset. All experiments are conducted using LLaMA-7B.

Table~\ref{tab:arxiv} presents the experimental results, demonstrating that \tool{} maintains its effectiveness when evaluated on different datasets, thereby confirming its generalizability across various data sources. Specifically, \tool{} exhibits consistent performance across temporal boundaries. For the Arxiv 18\&23 dataset, when $N_\text{unseen}=100$, \tool{} achieves 92.0\% accuracy with only 4 false positives and 36 false negatives, correctly identifying 64 true positives and 396 true negatives. This stability across different temporal splits demonstrates \tool{}'s robustness against temporal drift.

Moreover, when comparing performance across different temporal configurations (e.g., using 2018 versus 2022 data as member sets), \tool{} demonstrates remarkable consistency. The average accuracy remains stable at 83.8\% for the 2018 configuration and 84.1\% for the 2022 configuration, with minimal variation in false positive and false negative rates. These results provide strong evidence that \tool{} is resilient to temporal data distribution shifts and maintains reliable performance across different time periods.

\begin{table}[h]
    \centering
    \vspace{-3mm}
    \caption{EVALUATING THE EFFECTIVENESS OF \tool on ArxivMIA.}
     \vspace{-3mm}
    \resizebox{0.9\linewidth}{!}{
    \begin{tabular}{c|c|c|c|cccc}
    \hline
    
    \multicolumn{1}{c|}{LLaMA 7B} &\multicolumn{1}{c|}{$N_{\text{unseen}}$}& Acc.(\%) & bAcc.(\%)  & TP & FP & FN & TN \\
    \hline\hline
   \multirow{4}{*}{\rotatebox{90}{\scriptsize  Arxiv 18\&23}} &100  &92.0 &81.5&64&4&36&396   \\
   \multicolumn{1}{c|}{}&200&83.0 &78.8 &115&0&85&300  \\
   \multicolumn{1}{c|}{}&300&81.6&84.7 &208&0&92&200  \\
   \multicolumn{1}{c|}{}&400&78.6&86.6 &293&0&107&100 \\
 
   \hline

   \multirow{4}{*}{\rotatebox{90}{\scriptsize  Arxiv 22\&23}} &100  &91.2 &85.9&77&21&23&379   \\
   \multicolumn{1}{c|}{}&200&85.8&82.3 &129&0&71&300  \\
   \multicolumn{1}{c|}{}&300 &78.4&82.0&192&0&108&200  \\
   \multicolumn{1}{c|}{}&400&80.8&88.0 &304&0&96&100 \\
 
   \hline
    \end{tabular}
    }
    \label{tab:arxiv}
    \vspace{-3mm}
\end{table}

\begin{answerBox}
\vspace{-1mm}
\textbf{Answer to RQ3:} Experiment results show that \tool{} effectively identifies seen files across two LLaMA-based LLMs and one Transformer-based LLM, indicating that \tool{} can generalize well across different LLM architectures. Moreover, \tool's effectiveness persists on the ArxivMIA dataset, further demonstrating its generalizability across different data sources. 
\vspace{-1mm}
\end{answerBox}

\subsection{RQ4: Effectiveness in Black-Box Setting}




While our main experiments focus on the white-box setting, where \tool{} has full access to model parameters enabling uncertainty estimation methods (BLoB, MCD, and Ensemble), we also explore a black-box variant where only API-level log-probabilities can be queried without parameter access.

To perform uncertainty quantification in this black-box setting, we construct $N = 10$ semantically varied prompts for each text snippet, asking whether the model has "seen" the text before, with outputs restricted to binary responses (0 or 1). We then query the black-box model and collect the log-probabilities of tokens "0" and "1" across these prompts to derive uncertainty metrics, as described in \S\ref{metrics}. The calculated uncertainty metrics then undergo the same aggregation and clustering procedures as the white-box pipeline described in \S\ref{sec:seen_file_detection}.

\begin{table}[h]
    \centering
    \vspace{-4mm}
    \caption{Evaluating the effectiveness of \tool{} using GPT-3.5-Turbo-Instruct in detecting seen files. L stands for the length of WIKI sample.}
     \vspace{-2mm}
    \resizebox{0.9\linewidth}{!}{
    \begin{tabular}{cc|c|cccc}
    \hline
    \multicolumn{2}{c|}{WikiMIA 24} & \multicolumn{5}{c}{GPT 3.5 Turbo Instruct} \\ \hline
    \multicolumn{1}{c|}{L} & \multicolumn{1}{c|}{Tool} & Acc.(\%)  & TP & FP & FN & TN \\
    \hline\hline
    
   \multicolumn{1}{c|}{\multirow{3}{*}{32}}&\toolppl&50.1&776&775&3&4 \\
   \multicolumn{1}{c|}{}&\toolloss&52.3&538&503&241&276 \\
   \cline{2-7}
   \multicolumn{1}{c|}{} &\textbf{\tool{}}  & \textbf{64.6}  & 589 &362  & 190  &417   \\
   \hline
   
   \multicolumn{1}{c|}{\multirow{3}{*}{64}}&\toolppl&50.1&3&2&695&696 \\
   \multicolumn{1}{c|}{}&\toolloss&50.1&697&694&1&4 \\
   \cline{2-7}
   \multicolumn{1}{c|}{} &\textbf{\tool{}}  & \textbf{66.8} &515 &281& 183 &417 \\ 
   \hline
   
    \multicolumn{1}{c|}{\multirow{3}{*}{128}}&\toolppl&50.0&0&0&286&286\\
   \multicolumn{1}{c|}{}&\toolloss&50.0&0&0&286&286 \\
   \cline{2-7}
    \multicolumn{1}{c|}{} &\textbf{\tool{}}   & \textbf{66.3} & 197& 104 & 89 & 182 \\
   \hline
   
     \multicolumn{1}{c|}{\multirow{3}{*}{256}}&\toolppl&51.1&28&26&65&67 \\
   \multicolumn{1}{c|}{}&\toolloss&53.2&37&31&56&62 \\
   \cline{2-7}
   \multicolumn{1}{c|}{} &\textbf{\tool{}}  & \textbf{68.8} &67  & 32 & 26 & 61 \\ 
   \hline
    \end{tabular}
    }
    \label{black_box}
    \vspace{-3mm}
\end{table}

Table~\ref{black_box} summarizes the results under an API-only black-box setting using GPT-3.5-Turbo-Instruct. Despite the absence of internal access, \tool{} achieves up to 68.8\% accuracy on WikiMIA-24 dataset, which is available on Hugging Face under the repository “wjfu99/WikiMIA", outperforming probability-based baseline methods by more than 15 points. These findings demonstrate that uncertainty-based copyright detection remains feasible under API-only black-box access.

\begin{answerBox}
\vspace{-1mm}
\textbf{Answer to RQ4:} \tool{} remains effective under API-only black-box access, achieving up to 68.8\% accuracy, demonstrating that uncertainty-based copyright detection is feasible even without model parameter access.

\vspace{-2mm}
\end{answerBox}

\vspace{-2mm}
\section{Discussion}
\vspace{-1mm}
\label{sec:threats}

\noindent \textbf{File-level Detection.}
Our work operates at the file level, treating each file as either entirely seen or entirely unseen by the target LLM. This design choice aligns with established practices in file-level membership inference research~\cite{meeus2024did}.
While we ensure that only books published after the target LLM's release date are classified as unseen, we acknowledge the possibility of partial exposure through snippets within these files.
However, such quoted content from older sources constitutes only a minimal fraction (less than 5\%) of the total content in unseen files. To preserve data integrity, our approach maintains focus on file-level detection without removing these incidental quotes.
Future research could extend this framework by incorporating snippet-level uncertainty analysis to detect and characterize partial exposure patterns within individual files.

\noindent\textbf{Temporal shift of Dataset.} To further examine whether \tool’s uncertainty signal is affected by temporal or domain differences, we conduct additional experiments on the ArXiv dataset, where each configuration mixes 2023 papers (non-members) with member papers from nearby years (2018 or 2022). 
As shown in Table~\ref{tab:arxiv}, \tool achieves consistently high accuracy for both year settings, comparable to that on BookMIA. 
These results confirm that \tool captures genuine membership exposure rather than spurious time-dependent artifacts, demonstrating robustness across domains and publication periods.

\noindent\textbf{Computational Overhead.} \tool demonstrates computational efficiency in practice. Fine-tuning on the BookMIA dataset (100 books containing 10,000 samples with 300 tokens each) requires approximately 15 minutes for BLoB, 15 minutes for MCD, and 150 minutes for the 10-model ensemble approach. These training times demonstrate that \tool's uncertainty estimation methods impose minimal computational overhead while maintaining high detection accuracy.

\noindent\textbf{Black-Box LLM Applicability.} 
While our primary focus centers on white-box LLMs where model internals are accessible, we believe our approach can be adapted to black-box scenarios through API-based uncertainty estimation techniques~\cite{xiong2023can,tonolini-etal-2024-bayesian}.
Our exploration in RQ4 indicates that the methodology remains effective under black-box constraints. Future work could investigate these adaptations to extend \tool's applicability 
to a broader range of proprietary LLMs with limited internal access.
\vspace{-4mm}
\section{Conclusion}
\vspace{-2mm}
In this paper, we present \tool{}, a general framework for detecting whether copyrighted files were used in the training of a target LLM. Unlike traditional membership inference attacks that rely on shadow models, handcrafted thresholds, or large amounts of known training data, \tool{} leverages uncertainty signals as a reliable indicator. Our analysis demonstrates that uncertainty can effectively distinguish between seen and unseen snippets. Through strategic snippet segmentation and unsupervised clustering, \tool{} achieves consistently high accuracy and balanced accuracy on extensive experiments across LLM architectures. Overall, \tool{} represents the first uncertainty-based approach for file-level copyright detection in LLMs, providing a practical mechanism for auditing training data usage.
\vspace{-2mm}




\bibliographystyle{IEEEtran}
\vspace{-2mm}
\bibliography{cite}

\end{document}